%% file: iwpss21eoscsp.tex
\def\year{2021}\relax
\documentclass[letterpaper]{article} %
\usepackage{aaai21}  %
\usepackage{times}  %
\usepackage{helvet} %
\usepackage{courier}  %
\usepackage[hyphens]{url}  %
\usepackage{graphicx} %
\usepackage{natbib}  %
\usepackage{caption} %
\makeatletter
\def\copyright@text{Copyright \copyright\space \copyright@year,
All rights reserved.}
\makeatother

\usepackage{xcolor}
\usepackage{paralist}
\usepackage{caption}
\usepackage{subcaption}
\usepackage{tikz}
\usetikzlibrary{patterns}
\usetikzlibrary{fit}
\usepackage[linesnumbered,ruled]{algorithm2e}
\SetAlFnt{\small}
\SetAlgoSkip{}
\usepackage{amsmath}
\usepackage{amssymb}

\DeclareMathOperator*{\maximize}{\text{max}}
\DeclareMathOperator*{\argmax}{arg\,max}
 
\newcommand{\before}{\ensuremath{\beta}}

\newtheorem{definition}{Definition}
\usepackage{tikz}
\usetikzlibrary{decorations.pathmorphing, patterns,shapes}
\usetikzlibrary{arrows.meta}
\usepackage{svg}
\usetikzlibrary{calc}
\def\centerarc[#1](#2)(#3:#4:#5){\draw[#1] ($(#2)+({#5*cos(#3)},{#5*sin(#3)})$) arc (#3:#4:#5); }
\tikzset{
capture/.style={fill=gray!40,opacity=0.5,draw=gray}, 
communication/.style={fill=gray!10,opacity=0.5,decorate, decoration={snake, segment length=1mm, amplitude=0.3mm},draw=gray},
groundcomm/.style={line width=0.6mm,color=gray}}
\usepackage{tikz-3dplot}
\usetikzlibrary{shapes.misc}
\tikzset{cross/.style={cross out, draw=black, fill=none, minimum size=2*(#1-\pgflinewidth), inner sep=0pt, outer sep=0pt}, cross/.default={2pt}} %
\usetikzlibrary{patterns}
\usetikzlibrary{fit}
\definecolor{mygreen}{HTML}{8dd3c7}
\definecolor{myyellow}{HTML}{ffffb3}
\definecolor{myblue}{HTML}{bebada}
\tikzset{obs/.style={fill=myblue, draw=myblue!50, minimum width=0.5cm, minimum height=0.8cm}, 
aobs/.style={fill=mygreen, minimum width=0.5cm, draw=mygreen!50, minimum height=0.8cm, text=black},
bobs/.style={fill=myyellow, minimum width=0.5cm, draw=myyellow!50, minimum height=0.8cm, text=black},
aex/.style={pattern=north west lines, pattern color=mygreen,draw=white},
bex/.style={pattern=north west lines, pattern color=myyellow!70!orange,draw=white},
agent/.style={draw, circle, minimum width=1cm}}
\usetikzlibrary{positioning}
\usetikzlibrary{arrows.meta}

\newcommand{\only}[1]{}

\title{Auction-based and Distributed Optimization Approaches for Scheduling Observations in Satellite Constellations with Exclusive Orbit Portions}
\author{
    Gauthier Picard\\
}
\affiliations{
    ONERA/DTIS, Université de Toulouse\\
    gauthier.picard@onera.fr
}
\begin{document}

\maketitle

\begin{abstract}
We investigate the use of multi-agent allocation techniques on problems related to Earth observation scenarios with multiple users and satellites. We focus on the problem of coordinating users having reserved exclusive orbit portions and one central planner having several requests that may use some intervals of these exclusives. We define this problem as Earth Observation Satellite Constellation Scheduling Problem (EOSCSP) and map it to a Mixed Integer Linear Program. As to solve EOSCSP, we propose market-based techniques and a distributed problem solving technique based on Distributed Constraint Optimization (DCOP), where agents cooperate to allocate requests without sharing their own schedules. These contributions are experimentally evaluated on randomly generated EOSCSP instances based on real large-scale or highly conflicting observation order books.
\end{abstract}

\section{Introduction}
\label{sec:introduction}
\input{introduction.tex}

\section{EOSCSP Model}
\label{sec:definitions}
\input{definitions.tex}

\section{Centralized Problem Solving for EOSCSP}
\label{sec:centralized}
\input{milp.tex}

\section{Auction-based Coordination for EOSCSP}
\label{sec:auctions}
\input{auctions.tex}

\section{DCOP-based Coordination for EOSCSP}
\label{sec:dcop}
\input{dcop.tex}

\section{Experimental Evaluation}
\label{sec:experiments}
\input{experiments.tex}

\section{Conclusion and Synthesis}
\label{sec:conclusion}
\input{conclusion.tex}

\section*{Acknowledgments}

This work has been performed with the support of the French government in the context of the “Programme d’Invertissements d’Avenir”, namely by the BPI PSPC project “LiChIE”, coordinated by Airbus Defence and Space.

\bibliography{eoscsp,auctions}

\end{document}

%% file: introduction.tex
\begin{figure}
\resizebox{\columnwidth}{!}{
    \centering

    \tikzset{
capture/.style={fill=gray!40,opacity=0.5,draw=gray}, 
communication/.style={fill=gray!10,opacity=0.5,decorate, decoration=snake,draw=gray},
groundcomm/.style={line width=1mm,color=gray}}

\begin{tikzpicture}[xscale=0.8,yscale=1.5]
\sf
\LARGE

\begin{scope}
	\clip(-16,1) rectangle (16,11);
    \clip (90,-85) arc (0:180:90);
    \path[fill=gray!50] (-12,0) -- (-11,1) -- (-9, 2) -- (-8, 1.5) -- (-7, 3) -- (-8, 5) -- (-5, 5) -- (-4, 3) -- (-4,2) -- (-3, 2.5) -- (-1, 2) -- (-1, 4) -- (2, 3) -- (2, 2) --(4, 1) -- (7, 0.5) -- (8, 2) -- (7, 4) -- (10, 6) -- (11, 3) -- (11, 1) -- (13, 0.5) -- (14, 2) -- (17, 4) -- (16, 0);
\end{scope}
\begin{scope}
    \clip(-16,1) rectangle (16,11);
    \draw[line width=1mm] (90,-85) arc (0:180:90);
    \draw[dashed,gray] (95,-85) arc (0:180:95);
\end{scope}

\begin{scope}
    \clip(-16,1) rectangle (16,11);
	\path[communication] (-9.2, 5.2) -- (-13, 14) -- (-17, 14) -- (-17, 9) -- (-9.2, 5.2);
	\path[communication] (13.8, 5) -- (13, 14) -- (8, 14) -- (13.8, 5);

\node[] (cloud1) at (-4,7) {\rotatebox{0}{\includegraphics[width=2.5cm]{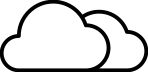}}};
\node[] (cloud2) at (-8,8) {\rotatebox{0}{\includegraphics[width=2.5cm]{cloud}}};
\node[] (cloud3) at (3,7) {\rotatebox{0}{\includegraphics[width=2.5cm]{cloud}}};
\node[] (cloud4) at (5,8) {\rotatebox{0}{\includegraphics[width=2.5cm]{cloud}}};
\node[] (cloud5) at (-4,7) {\rotatebox{0}{\includegraphics[width=2.5cm]{cloud}}};

\path[capture] (-12.75, 8.5) -- (-12, 3.25) -- (-11, 3.5) -- (-12.75, 8.5);
\path[capture] (0.5, 9.5) -- (cloud3.west) -- (cloud3.north) -- (0.5, 9.5);
\path[capture] (12.5,8.75) -- (6, 4) -- (5.5, 4.5) -- (12.5,8.75);

\node[inner sep=0pt,label={[label distance=-0.2cm]270:Mission center $u_0$}] (station1) at (-13.4,2.2) {\includegraphics[width=2cm]{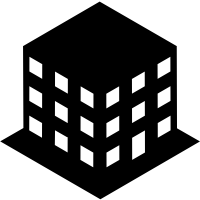}};
\node[inner sep=0pt,label={[label distance=0cm]270:Ex. User $u_1$}] (agency1) at (-3,4.5) {\includegraphics[width=2cm]{building}};
\draw[groundcomm] (station1) to[bend right=0] (agency1);
\node[inner sep=0pt,label={[label distance=0cm]270:Agency}] (agency2) at (6,2.5) {\includegraphics[width=2cm]{building}};
\node[inner sep=0pt,label={[label distance=0cm]270:Comm. station}] (comm1) at (-8.5,4.5) {\scalebox{-1}[1]{\includegraphics[width=1.5cm]{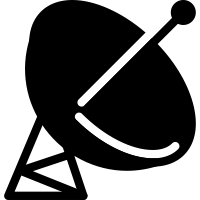}}};
\node[inner sep=0pt,label={[label distance=0cm]270:Ex. User $u_2$}] (comm2) at (13.75,3.5) {\rotatebox{0}{\scalebox{-1}[1]{\includegraphics[width=1.5cm]{building}}}};
\node[inner sep=0pt] (subcomm1) at (13.75,4.2) {\rotatebox{-35}{\scalebox{-1}[1]{\includegraphics[width=1cm]{antenna}}}};
\node[inner sep=0pt, above= -0.5cm of agency1] (subcomm2) {\rotatebox{-35}{\scalebox{-1}[1]{\includegraphics[width=1cm]{antenna}}}};
\node[inner sep=0pt,label={[label distance=-0.3cm]90:EO Satellite 1}] (satellite1) at (-12.9,9.2) {\rotatebox{-25}{\includegraphics[width=2cm]{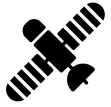}}};
\node[inner sep=0pt,label={[label distance=-0.3cm]90:EO Satellite 2}] (satellite2) at (0,10) {\rotatebox{5}{\includegraphics[width=2cm]{satellite}}};
\node[inner sep=0pt,label={[label distance=-0.0cm]90:EO Satellite 3}] (satellite3) at (13,9.2) {\rotatebox{255}{\includegraphics[width=2cm]{satellite}}};

\path[communication] (subcomm2.north) -- (-6, 14) -- (-2, 14) -- (subcomm2.north);

\draw[groundcomm] (satellite1) to[bend left = 0] (comm1);

\draw[groundcomm] (station1) to[bend right=10] (agency2);
\draw[groundcomm] (agency1) to[bend left=10] (agency2);
\draw[groundcomm] (station1) to[bend left=15] (comm1);
\draw[groundcomm] (agency2) to[bend left=0] (comm2);
\draw[groundcomm] (satellite3) to[bend left = 0] (subcomm1);
\draw[groundcomm] (station1) to[bend right=5] (-16, 3);
\draw[groundcomm] (comm2) to[bend right=5] (16, 3);

\end{scope}
\end{tikzpicture}
}
\caption{An Earth Observation system composed of a main mission center $u_0$, distributed stations (with ranges), agencies emitting observation requests (to mission center), EOS (with image footprint), communication satellites (linking EOS), and exclusive users with their own ground stations.}
\label{fig:system}
\end{figure}

Recent years have shown a large increase in the development of satellite constellations. Instead of considering individual satellites, they take advantage of a group of satellites, some of them often sharing the same orbital planes, to provide richer services like positioning, telecommunication or Earth observation \cite{Walker1984}. With few satellites in a constellation (\textit{e.g.} two in the PLEIADES project \cite{Lemaitre2002}), and in low or medium Earth orbits (altitude inferior to 35,000km), no region on Earth is permanently covered by the constellation at any time. So, the main motivation to increase the size of these constellations is to allow to capture with a high reactivity any point on Earth, as the Planet company is doing with more than 150 Earth Observation Satellites (EOS) \cite{Shah2019}. But, operating numerous EOS requires improving cooperation between the assets and on-board autonomy in order to make the best use of the system, which becomes a highly combinatorial task. Besides their growing number, constellations' composition is evolving too. Recent technological advances allow the production and deployment of agile EOS able to change their orientation, and to provide multiple types of image shooting with multiple sensors. While providing richer services to multiple users, this adds many degrees of freedom and decision variables to schedule EOS activity, and opens many challenges \cite{Wang2020}. 
Among these challenges, we focus on the collective scheduling of observations on a set of satellites on which some users have \textit{exclusive access to some orbit portions}, using distributed techniques, as to spread decisions among the different users of the constellation. %
This anwers to strong user expectations to benefit on the one hand from the advantages of a system shared between several stakeholders (reduction of costs compared to a very expensive global system) and on the other hand from the advantages of a proprietary system (ability to do what one wants with the satellite and potentially without disclosing it to others). While the literature about multi-satellite scheduling is rich, as confirmed by a recent review paper \cite{Wang2020}, considering satellite constellations as shared resources requiring multiple users to coordinate as to allocate tasks within exclusive orbit portions is a completely novel problem, we address in this paper, as illustrated in Figure~\ref{fig:system}.

In \cite{Phillips2021}, marked-based approaches are proposed to allocate observation tasks to a set of satellite, where each satellite is managed by a different mission center. Mission centers coordinate their allocation using auction-based protocols, by bidding on the open observations depending on the impact on the on-board plan and its reward (valued using the incidence angle of the scheduled observations). Contrary to this approach, in this study, the distribution is related to some exclusive users having full control on some orbit portions (using full direct tasking operation) or having bought some orbit portions outside direct communication, on which they have full priority to schedule observations. Here, the fact that schedules cannot be performed by a single authority, for privacy reason in exclusive windows, is a strong requirement. This is the reason to provide distributed scheduler where agents coordinate without disclosing their plans, while meeting coupling constraints like satellite capacity or inter-observation configuration time, that could not be guaranteed by non-coordinated schemes where users make their plans in parallel. We will investigate here two different distributed resource allocation and coordination schemes: market-based and DCOP-based (distributed constraint optimization).

Section~\ref{sec:definitions} illustrates and defines Earth Observation Satellite Constellation Scheduling Problem (EOSCSP). Section~\ref{sec:centralized} focuses on centralized solution methods Mixed-Integer Linear Program (MILP) and greedy approach to EOSCSP. Section~\ref{sec:auctions} expounds some market-based approaches to solve EOSCSP, using different auction schemes (PSI, SSI and CBBA), while Section~\ref{sec:dcop} proposes another approach to coordination between exclusive users using distributed constraint optimization (DCOP). 

We experimentally evaluate these different algorithms using randomly generated instances, in Section~\ref{sec:experiments}. Finally, Section~\ref{sec:conclusion} concludes the paper with some perspectives.

%% file: definitions.tex
This section illustrates the problem we investigate using a sample scenario, and then provides some core definitions.

\subsection{Sample Scenario}

\begin{figure}[tb]
\begin{center}
\includegraphics[width=\columnwidth]{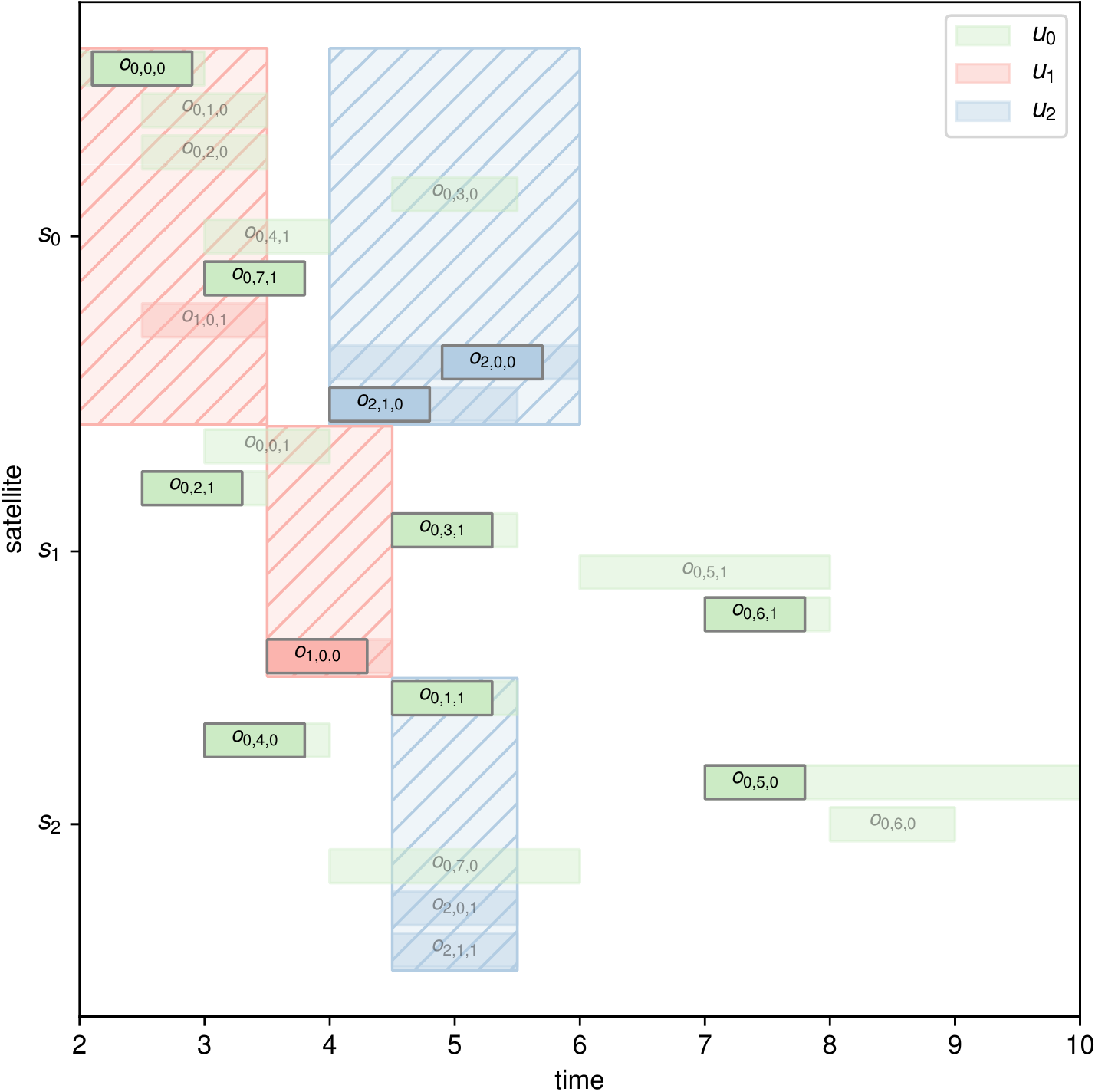}
\end{center}
  \vspace{-1em}\caption{An example with 3 satellites, 2 exclusive users (red and blue) with exclusives (hashed areas), and 1 non-exclusive user (green). Observation time windows appear as transparent surfaces. Solid surfaces represent an optimal solution.\label{fig:scenario}}
\end{figure}

Figure~\ref{fig:scenario} illustrates a scenario, where we consider:
\begin{inparaitem}[]
\item 3 satellites, each having a given planning period (e.g. planning on the next orbit, or on horizons depending on the communication windows between the satellite and ground stations);
\item 1 user $u_0$ without exclusive orbit portion;
\item 2 users having exclusive orbit portions such that
  \begin{inparaitem}[]
  \item user $u_1$ owns exclusives on satellite $s_0$ and on satellite $s_1$ (hashed red),
  \item user $u_2$ owns exclusives on satellite $s_0$ and on satellite $s_2$ (hashed blue);
  \end{inparaitem}
\item several requests to be performed before a due date, denoted $r_{i,j}$ for the $j$th request for user $i$;
\item several observation opportunities (simply observations) per request, denoted $o_{i,j,k}$ for the $k$th observation for the $j$th request of the $i$th user. Only one observation should be planned to fulfill the request on temporal slots depending on the satellites' orbits and the position of zones of interest (slots are represented as transparent areas). More precisely, we consider 2 observations per request, such that
  \begin{inparaitem}[]
  \item observations $o_{1,0,0}$ and $o_{1,0,1}$ are private for user $u_1$ (in red),
  \item observations $o_{2,0,0}$, $o_{2,0,1}$, $o_{2,1,0}$, and $o_{2,1,1}$ are private for user $u_2$ (in blue),
  \item observations $o_{0, j, k}$'s (in green) which are directly requested to the central scheduler $u_0$ by other clients without exclusives. The proposed solution fulfills all requests, by allowing non exclusive user $u_0$ to position observation on exclusive orbit portions (e.g. $o_{0,0,0}$ in on $u_1$'s exclusive on satellite $s_0$).
\end{inparaitem}
  \item A simplified energy constraint states that a satellite cannot perform more than $n_\textrm{max}$ observations on its scheduling period (here, $n_\textrm{max}=4$),
  \item and there are minimal transition times between two observations $o$ and $p$, depending on $o$ and $p$ and the date at which the transition is triggered on a given satellite.
\end{inparaitem}
At the global level, each exclusive user ($u_1$ or $u_2$) could have its own scheduling system to manage its exclusive periods, and a central scheduling system ($u_0$) manages observations $o_{0,j,k}$'s. In the end, every user and the central scheduler have a local scheduling problem to solve. Solving them in a separate manner may lead the central scheduler not to be able to book slots on exclusive orbit portions, while it may improve the solution. Without coordination, and with a non-cooperative management of exclusive slots, the overall schedule might not be optimal, wrt the number of possible scheduled observations. Moreover, exclusive users may gain from this cooperation by making profits from observation scheduled on their orbit portions. Thus, we propose here to coordinate the scheduling processes between users. %

\definecolor{mygreen}{HTML}{8dd3c7}
\definecolor{myyellow}{HTML}{ffffb3}
\definecolor{myblue}{HTML}{bebada}
\tikzset{obs/.style={fill=myblue, draw=myblue!50, minimum width=0.5cm, minimum height=0.8cm}, 
aobs/.style={fill=mygreen, minimum width=0.5cm, draw=mygreen!50, minimum height=0.8cm, text=black},
bobs/.style={fill=myyellow, minimum width=0.5cm, draw=myyellow!50, minimum height=0.8cm, text=black},
aex/.style={pattern=north west lines, pattern color=mygreen,draw=white},
bex/.style={pattern=north west lines, pattern color=myyellow!70!orange,draw=white}}

\subsection{Definitions and Notations}

Let's provide the core concepts of this scheduling problem. 

\begin{definition}
An \emph{Earth Observation Satellite Constellation Scheduling with Exclusives Problem} (or EOSCSP) is defined by a tuple $P = \langle\mathcal{S},\mathcal{U},\mathcal{R},\mathcal{O}\rangle$, such that $\mathcal{S}$ is a set of satellites, $\mathcal{U}$ is a set of users, $\mathcal{R}$ is a set of requests, and $\mathcal{O}$ is a set of observations to schedule to fulfill requests in $\mathcal{R}$.
\end{definition}

\begin{definition}A \emph{satellite} is defined as a tuple $s=\langle t_s^\text{start}, t_s^\text{end}, \kappa_s, \tau_{s}\rangle$ with $t^\text{start}_s\in\mathbb{R}$ the start time of its orbit plan, $t^\text{end}_s\in\mathbb{R}$ the end time of its orbit plan, $\kappa_s\in\mathbb{N}^+$ its capacity (i.e. the maximum number of observations during its orbit plan), $\tau_{s}: \mathcal{O}\times\mathcal{O}\rightarrow\mathbb{R}$ the function defining transition times between two given observations.
\end{definition}

\begin{definition}A \emph{user} is defined as a tuple $u=\langle e_u, p_u\rangle$ with a (possibly empty) set of exclusive time windows $e_u = \{(s,(t^\text{start}, t^\text{end}))\ |\ s\in\mathcal{S}, [t^\text{start}, t^\text{end}] \subseteq [t_s^\text{start}, t_s^\text{end}])\}\subset (\mathcal{S}\times(\mathbb{R}\times\mathbb{R}))$, and a priority $p_u\in\mathbb{N}^+$ (the lower the better, used in case of conflict). We note $\mathcal{U}^\mathsf{ex}$ (resp. $\mathcal{U}^\mathsf{nex}$) the set of users owning (resp. not owning) exclusives.
\end{definition}

We assume here that only one user has no exclusive orbit portion, the central planner, denoted $u_0$, i.e. $\mathcal{U}^\textsf{nex} = \{u_0\}$, and there is no overlapping exclusive portions.

\begin{definition}A \emph{request} is defined as a tuple $r=\langle t_r^\text{start}, t_r^\text{end}, \Delta_r, \rho_r, p_r, u_r, \theta_r\rangle$, with a validity time window defined by $t_r^\text{start}\in\mathbb{R}$ and $t_r^\text{end}\in\mathbb{R}$, a duration $\Delta_r\in\mathbb{R}$, a reward $\rho_r\in\mathbb{R}$ if $r$ is fulfilled, a latitude-longitude-altitude position (LLA) to observe $p_r$, a requester $u_r\in\mathcal{U}$ and a list $\theta_r\in 2^\mathcal{O}$ of observation opportunities to fulfill the request. 
\end{definition}

$\theta_r$ is dynamically computed on current constellation configuration and requested LLA position $p_r$, since several agile satellites, by changing their orientation may acquire the same position, thus generating several observation opportunities.

\begin{definition}An \emph{observation} is defined as a tuple $o=\langle t_o^\text{start}, t_o^\text{end}, \Delta_o, r_o, \rho_o, s_o, u_o, p_o\rangle$, with a validity time window defined by $t_o^\text{start}\in\mathbb{R}$ and $t_o^\text{end}\in\mathbb{R}$, a request $r_o$ to which it contributes, a duration $\Delta_o\in\mathbb{R}$ ($\Delta_o = \Delta_{r_o}$), a reward $\rho_o\in\mathbb{R}$ (combined from $r_o$ and information about the weather), a satellite $s_o$ on which this observation can be scheduled, an owner $u_o\in\mathcal{U}$ ($u_o = u_{r_o}$), and a priority $p_o\in\mathbb{N}^+$ ($p_o = p_{r_o}$). 
\end{definition}

The difference between request reward and observation reward comes from the fact that, in practice, weather conditions or incidence angle of an observation may increase or decrease the basic reward for a given request. So, our model can consider different rewards, but in this study we only focus on cases where observation rewards are directly inherited from the requests.

\begin{definition}
A \emph{solution} to an EOSCSP is a mapping $\mathcal{M} = \{(o,t)\ |\ o\in\mathcal{O}, t\in[t_o^\text{start}, t_o^\text{end}]\}$ assigning a start time to at most one observation per request such that exclusive users have their observations scheduled on their respective exclusive windows, and the overall reward is maximized (sum of the rewards of the scheduled observations): $\argmax_{\mathcal{M}}\sum_{(o,t)\in\mathcal{M}} r_o$.
\end{definition}

\begin{definition}
An \emph{EOSCSP for user $u$}, denoted $P[u] = \langle\mathcal{S},\mathcal{U},\mathcal{R}[u], \mathcal{O}[u]\rangle$ (or EOSCSP$[u]$), is an EOSCSP, sub-problem of another EOSCSP $P = \langle\mathcal{S},\mathcal{U},\mathcal{R},\mathcal{O}\rangle$ restricted to requests and observations from $u$, where $\mathcal{R}[u] = \{r |\ r\in\mathcal{R}, u_r = u\}\subseteq\mathcal{R}$ and $\mathcal{O}[u] = \{o |\ o\in\mathcal{O}, u_o = u\}\subseteq\mathcal{O}$.
\end{definition}

More generally, we note $P[x]$ (resp. $P[x]$) the problem $P$ limited to the only components related to $x$, $x$ being a request, an observation or a satellite. 
Later on, we will also use the notations $P[\emptyset|\mathcal{M}]$ (resp. $P[u_{l},\ldots, u_{m}|\mathcal{M}]$) to define the problem (resp. sub-problem for users $u_{l},\ldots, u_{m}$) given some predefined allocation $\mathcal{M}$ of some observations. Moreover, we will use notation $\overline{P}$ to appoint the EOSCSP $P$, where only requests and related observations that can be scheduled outside exclusive are considered (i.e. observations whose time windows intersect non exclusive orbit portions). Finally, we note the union of two problems $P = \langle\mathcal{S},\mathcal{U},\mathcal{R},\mathcal{O}\rangle$ and $P' = \langle\mathcal{S}',\mathcal{U}',\mathcal{R}',\mathcal{O}'\rangle$, $P \cup P' = \langle\mathcal{S}\cup\mathcal{S}',\mathcal{U}\cup\mathcal{U}',\mathcal{R}\cup\mathcal{R}',\mathcal{O}\cup\mathcal{O}'\rangle$.

%% file: milp.tex
We present here centralized approaches to EOSCSP. 
First, this planning problem is modeled as a MILP. 
Decision variables are the following.
\begin{inparaitem}[]
\item $x_{s,o}\in\{0,1\}$ is the decision to perform observation $o$ from satellite $s$,
\item $t_{s,o}\in\mathbb{R}$ is the start date for the observation $o$ on satellite $s$, 
\item $\before_{s,o,p} \in \{0,1\}$ is the precedence between observations on the same satellite, equals to $1$ if $o$ is before $p$ on $s$. 
\end{inparaitem}

\allowdisplaybreaks
\noindent\begin{align}
\smaller
\maximize_{x_{s,o}}\quad & \textstyle\sum_{o\in\mathcal{O}, s\in\mathcal{S}}\rho_o x_{s,o}\label{eq:milp-obj}\\
\text{s.t.}\quad & \forall s\in\mathcal{S}, \forall r \in \mathcal{R}, \forall o \in\mathcal{O}, \forall p\in\mathcal{O}, o\neq p\nonumber\\
& 2 - \before_{s,o,p} - \before_{s,p,o} \geq x_{s,o}\label{eq:milp-C1}\\
& 2 - \before_{s,o,p} - \before_{s,p,o} \geq x_{s,p}\label{eq:milp-C2}\\
& \beta_{s,o,p} + \beta_{s,p,o} \leq 1\label{eq:milp-C3b}\\
& t_{s,p} - t_{s,o} \geq \tau_{s}(o,p) + \Delta_o - \Delta_{s,o,p}^\text{max}\before_{s,o,p},\ \Delta_{s,o,p}^\text{max} > 0\label{eq:milp-C4}\\
& t_{s,o} - t_{s,p} \geq \tau_{s}(p,o) + \Delta_p - \Delta_{s,p,o}^\text{max}\before_{s,p,o},\ \Delta_{s,p,o}^\text{max} > 0\label{eq:milp-C5}\\
& \textstyle\sum_{o\in\mathcal{O}} x_{s,o} \leq \kappa_s\label{eq:milp-C6}\\
& \textstyle\sum_{o\in\theta(r)} x_{s,o} \leq 1\label{eq:milp-C7}\\
& x_{s,o} \in\{0,1\}\label{eq:milp-C8}\\
& t_{s,o} \in[t_o^{\text{start}},t_o^{\text{end}}] \subset\mathbb{R}\label{eq:milp-C9}\\
& \before_{s,o, p} \in\{0,1\}\label{eq:milp-C10}\\
\text{with}\quad  & \Delta_{s,o,p}^\text{max} = t_o^\text{end} - t_p^\text{start} + \Delta_o + \tau^s(o, p)\nonumber
\end{align}
(\ref{eq:milp-C1}) to (\ref{eq:milp-C5}) ensure precedence of observations and their distance is at least the transition time required on their satellite. (\ref{eq:milp-C6}) enforces the number of observations booked on a satellite does not exceed its capacity. (\ref{eq:milp-C7}) checks at most one observation per request is scheduled. (\ref{eq:milp-C8}) to (\ref{eq:milp-C10}) are domain definitions. 
This MILP can be solved using off-the-shelf solvers like CPLEX or Gurobi, but they will hardly scale up when dealing with larger problems (e.g. more than 100 observations with 3 satellites and 3  users). 
To ensure observations from exclusive users have priority over non-exclusive users' observations, their reward must be set to a high value. Thus, the solver will prefer scheduling exclusive observations within their time window instead of scheduling another observation with less priority.
While the solution to this problem is optimal, it requires each exclusive user to \textit{fully disclose request information} to the central planner.

As to solve large problems, one approach is to apply a greedy allocation consisting in planning first exclusive users's observations and then more urgent observations, as described in Algorithm~\ref{alg:greedy}. In practice, this is the technique used by most satellite/constellation operators and is a candidate competitor for benchmarking solution methods \cite{Cho2018,Wang2020}. For doing so, observations are sorted in increasing order on priority and start time criteria (line 2). Then, for each observation in this sorted list, the first free slot on its satellite orbit plan is found (line 4-8). This algorithm is not optimal, but provides very fast solutions. However, as for MILP, this solution requires sharing all the constraints and information with a central planner.

\DontPrintSemicolon
\SetKwProg{Fn}{Function}{}{}
\SetKwFunction{fs}{first\_slot}
\SetAlgoVlined
\vspace{-0.6em}
\begin{algorithm}
  	\caption{Greedy EOSCSP solver} \label{alg:greedy}
  	\KwData{An EOSCSP $P = \langle\mathcal{S}, \mathcal{U}, \mathcal{R}, \mathcal{O}\rangle$}
  	\KwResult{An assignment $\mathcal{M}$} %
	$\mathcal{M}\leftarrow \{\}$\;
	$\mathcal{O}^\text{sorted} \leftarrow \mathtt{sort}(\mathcal{O})$\;
	$R\leftarrow\{(s,[])\}\ |\ s\in\mathcal{S}\}$\;
	\For{$o\in\mathcal{O}^\text{sorted}$}{
		$t \leftarrow \mathtt{first\_slot}(o, P, R)$\;
		\If{$t\neq\emptyset$}{
			$\mathcal{M}\leftarrow \mathcal{M}\cup\{(o,t)\}$\;
			$\mathcal{O}^\text{sorted} \leftarrow \mathcal{O}^\text{sorted} \setminus\theta(r_o)$
		}
	}
	\KwRet{$\mathcal{M}$}
	~\nonumber\newline\newline
\Fn{\fs($o$, $P = \langle\mathcal{S}, \mathcal{U}, \mathcal{R}, \mathcal{O}\rangle$, $R$)}{
	\For{$(s, [t^\text{start},t^\text{end}]) \in \mathtt{domains}(o)$}{
		\If{$|R[s]| < \kappa_s$}{
			\eIf{$R[s] = []$}{
				\If{$t^\text{end} \geq t^\text{start} + \Delta_o$}{
					$R[s] = \{(o,(s, t^\text{start}))\}$\;
					\KwRet{$(s, t^\text{start})$}
				}
			}{
			$i\leftarrow 0$\;
			\While{$i \leq |R[s]|$}{
				$t^{\text{start}\prime}\leftarrow t^\text{start}$\;
				\If{$i > 0$}{
					$(o_{i-1}, (s, t_{i-1}))\leftarrow R[s][i-1]$\;
					$t^{\text{start}\prime}\leftarrow\max(t^\text{start}, t_{i-1} + \Delta_{o_{i-1}} + \tau_{s,o_{i-1},o})$
				}
				\If{$t^{\text{start}\prime} + \Delta_o \leq t^\text{end}$}{
					\eIf{$i = |R[s]|$}{
						$t^\text{upper}\leftarrow	t^\text{end}$\;
						$t^{\text{end}\prime}\leftarrow t^{\text{start}\prime} + \Delta_o$			
					}{
						$(o_i, (s, t_i))\leftarrow R[s][i]$\;
						$t^\text{upper}\leftarrow	t_i$\;
						$t^{\text{end}\prime}\leftarrow t^{\text{start}\prime} + \Delta_o + \tau_{s,o,o_i}$					
					}
					\If{$t^{\text{start}\prime} < t^{\text{end}\prime} \leq t^\text{upper}$}{
						$R[s] = \texttt{insert}(R[s], (o,(s, t^{\text{start}\prime})), i)$\;
						\KwRet{$(s, t^{\text{start}\prime})$}				
					}		
				}
				$i++$
			}
			}
		}
	}
	\KwRet{$\emptyset$}
	}
\end{algorithm}

%% file: auctions.tex
One vision to allocate resources and/or tasks between several agents (here, our exclusive users) consists in market-based approaches, that have proven their flexibility, efficiency, fairness, and privacy-preservation of users' plans and resources. In multi-robot task allocation problems, such approaches are used to allocate tasks to robots, and integrate them into their plans \cite{Dias2006}. In our setting, one could consider allocating requests to satellites by such market-based mechanisms, as proposed in \cite{Phillips2021}, with the difference that distribution is not related to satellites, but to exclusive users and their exclusive orbit portions. 
This is the approach we follow in this section. But first, let's introduce the auction-based mechanisms we will implement.  

\subsection{Some Background on Market-based Allocation}

\begin{figure}
\resizebox{\columnwidth}{!}{
\begin{tikzpicture}[align=center,node distance=2.75cm]
\sf
\node[agent] (auctioneer) at (0, 0) {$a$};
\node[above= 1cm of auctioneer] (phantom){}; 
\only<1->{
	\node[below = 2cm of auctioneer] (mid) {$\ldots$};
	\node[agent, below left= of auctioneer] (bidder2) {$b_2$}; 
	\node[agent, left= of auctioneer] (bidder1) {$b_1$}; 
	\node[agent, below right= of auctioneer] (bidder3) {$b_{n-1}$}; 
	\node[agent, right= of auctioneer] (bidder4) {$b_n$}; 
}
\only<2>{
	\foreach \i in {1,...,4} {
		\path[->,> = stealth] (auctioneer) edge[bend right=15] node[above,sloped]{\scriptsize 1: announcement} (bidder\i);
	}
}
\only<3>{
	\foreach \i in {1,...,4} {
		\path[->,> = stealth] (bidder\i) edge[loop below] node[below]{\scriptsize 2: valuation} (bidder\i);
	}
}
\only<4>{
	\foreach \i in {1,...,4} {
		\path[->,> = stealth] (bidder\i) edge node[above,sloped]{\scriptsize 3: bid} (auctioneer);
	}
}
\only<5>{
	\path[->] (auctioneer) edge[loop above] node[above]{\scriptsize 4: WDP} (auctioneer);
}
\only<6>{
	\foreach \i in {1,...,4} {
		\path[->,> = stealth] (auctioneer) edge[bend left=15] node[above,sloped]{\scriptsize 5: allocation} (bidder\i);
	}
}
\end{tikzpicture}
}
\caption{A sample auction process with one auctioneer $a$ and $n$ bidders $b_i$, following five main steps: \begin{inparaenum}[(1)]\item announcement of the items to allocate, \item valuation of the items or bundles by each bidder, \item communication of the computed bids, \item winner determination problem solving, and \item allocation of items to bidders.\end{inparaenum}}
\label{fig:auction}
\end{figure}
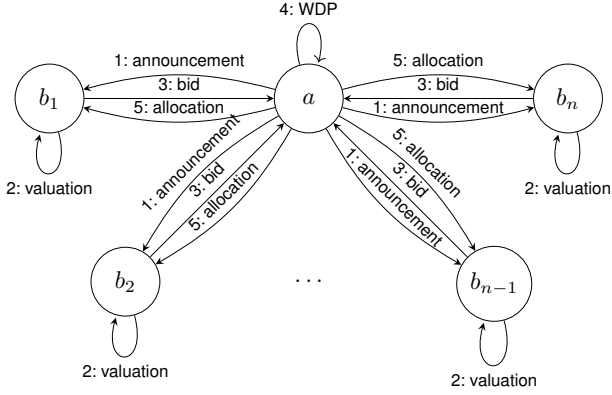

A generic task allocation framework consists in a set of resources and a set of tasks to be performed by resources. The objective is to assign tasks to resources so that it maximizes some objective (e.g. the number of assigned tasks, or the sum of the rewards of the tasks). So this is classical allocation problem that can be modeled as a MILP, as seen in previous section. Now, the idea is that the requests to be scheduled are open for bidding by an \textit{auctioneer}. \textit{Bidders} (the exclusive users) valuate the requests depending on their current plan, and bid for some requests, as illustrated in Figure~\ref{fig:auction}. The most expensive computations in this process are the bidding step by each bidder, which can have an exponential number of bundles to valuate, and the winner determination problem (WDP) which amounts to solving an Integer Linear Program with a potentially exponential size, and falls into the combinatorial auction (CA) frameworks \cite{Cramton2010}.

According to literature on multi-robot task allocation \cite{Dias2006} and multi-satellite observation allocation \cite{Phillips2021}, to overcome these computational limits, the classical relaxation consists in only allowing bidding on item (and not on bundles). When bidders bid on the whole set of items in parallel, we fall into PSI framework \cite{Koenig2006}. When the auctioneer announces items iterativelly, and bidders build their bid knowing the previous item allocation, we fall into the SSI framework \cite{Koenig2006}. In general PSI has very good performances with very limited computation time, while PSI solution quality are often limited, since bidders cannot easily reason on bundles. More recently, consensus-based bundle algorithm (CBBA) combines ideas from auctions and consensus to converge faster than SSI while yielding similar solutions and having the benefits of traditional consensus algorithms \cite{Choi2009}. CBBA is a fully distributed solution to implement
a computationally cheap variant of combinatorial auctions (CA).
Each bidder constructs a unique bundle of items it wishes to be
assigned to, with respect to the marginal cost associated with
the inclusion of the considered item into its current bundle.
Then during the consensus phase, the bidders compare their
bids with their teammates bids. If a robot is
outbid on an item $t$, it drops the item and all
the items added after it, as the exclusion of $t$ made the valuation
of their marginal cost obsolete. This algorithm 
have been extensively studied and modified to improve its
performances and adapt it to specific scenarios, like multi-satellite observation allocation \cite{Phillips2021}.

\subsection{Mapping EOSCSP to Auctions}

Mapping an EOSCSP $P$ to a market-based allocation problem is quite straightforward. Bidders are exclusives users in $\mathcal{U}^\textsf{ex}$, and items are non exclusive requests in $\mathcal{R}$ emitted by the central planner $u_0$, playing the role of auctioneer. The idea is that each exclusive user $u$ computes an initial plan $\mathcal{M}_u$ with its own requests by solving $P[u]$. Then, $u_0$ announces the requests, either as a whole (for PSI and CBBA) or iterativelly (for SSI). Each exclusive user $u$ valuates each single request with function \texttt{bid} (for PSI and SSI) or bundle with function \texttt{bundle} (for CBBA) by computing the marginal cost to integrate the given item or bundle $x$ in its current plan. We redirect the reader to the original CBBA paper for more details about the bundle construction \cite{Choi2009}. $\texttt{bid}(r, \mathcal{M}_u)$ simply amounts to solve $P[u] \cup P[r]$ and to assess the difference with the current plan $\mathcal{M}_u$. It returns the bid itself $\mathcal{B}_u[r]$ (best marginal cost) and the schedule for one observation to fulfill $r$, $\sigma_u[r] = (o, t)$. The bids (on single items or bundles) are then sent to the auctioneer $u_0$ (for PSI and SSI) which determines the winners, or to the other bidders sharing interest on the same request, namely $\mathcal{N}_u$, to find a consensus (for CBBA). Once the winners are determined, requests are allocated to the winners. If there remain some non allocated requests, $u_0$ attempts to schedule them outside any exclusive window. These processes are sketched in Algorithms~\ref{alg:psi}, \ref{alg:ssi} and \ref{alg:cbba}. 

\DontPrintSemicolon
\SetKwProg{Fn}{Function}{}{}
\SetAlgoVlined
\SetKwFor{ForEach}{for each}{do}{endfch}
\begin{algorithm}
  	\caption{\textsf{psi} EOSCSP solver} \label{alg:psi}
  	\KwData{An EOSCSP $P = \langle\mathcal{S}, \mathcal{U}, \mathcal{R}, \mathcal{O}\rangle$}
  	\KwResult{An assignment $\mathcal{M}$} %
	$\mathcal{M}_{u_0}\leftarrow \emptyset$\;
	\SetKwFor{ForEach}{for each}{do concurrently}{endfch}
	\ForEach{$u\in\mathcal{U}^\textsf{ex}$}{
		$\mathcal{M}_u\leftarrow \texttt{solve}(P[u])$\;
		\SetKwFor{ForEach}{for each}{do}{endfch}
		\lForEach{$r\in\mathcal{R}$}{
			$\mathcal{B}_u[r], \sigma_u[r] \leftarrow \texttt{bid}(r,\mathcal{M}_u)$
		}
		\tcp{send $\mathcal{B}_u, \sigma_u$ to $u_0$}
	}
	\SetKwFor{ForEach}{for each}{do}{endfch}
	\ForEach{$r\in\mathcal{R}$}{
		$w \leftarrow \argmax_{u\in\mathcal{U}^\textsf{ex}}\{\mathcal{B}_u[r]\}$\;
		$\mathcal{M}_{u_0} \leftarrow \mathcal{M}_{u_0} \cup\{\sigma_w[r]\}$\;
		$\mathcal{M}_w\leftarrow \mathcal{M}_w\oplus \sigma_w[r]$\hfill\tcp{send $\mathcal{M}_w[r]$ to $w$}
	}
	$\mathcal{M}_{u_0}\leftarrow \texttt{solve}(\overline{P[u_0|\mathcal{M}_{u_0}]})$\;
	\KwRet{$\bigcup_{u\in\mathcal{U}}{\mathcal{M}_u}$}
\end{algorithm}

\DontPrintSemicolon
\SetKwProg{Fn}{Function}{}{}
\SetAlgoVlined
\SetKwFor{ForEach}{for each}{do}{endfch}
\begin{algorithm}
  	\caption{\textsf{ssi} EOSCSP solver} \label{alg:ssi}
  	\KwData{An EOSCSP $P = \langle\mathcal{S}, \mathcal{U}, \mathcal{R}, \mathcal{O}\rangle$}
  	\KwResult{An assignment $\mathcal{M}$} %
	$\mathcal{M}_{u_0}\leftarrow \emptyset$\;
	\SetKwFor{ForEach}{for each}{do concurrently}{endfch}
	\lForEach{$u\in\mathcal{U}^\textsf{ex}$}{$\mathcal{M}_u\leftarrow \texttt{solve}(P[u])$}
	\SetKwFor{ForEach}{for each}{do}{endfch}
	\ForEach{$r\in\texttt{sorted}(\mathcal{R})$}{
		\ForEach{$u\in\mathcal{U}^\textsf{ex}$}{
			$\mathcal{B}_u[r], \sigma_u[r] \leftarrow \texttt{bid}(r,\mathcal{M}_u)$\;
			\tcp{send $\mathcal{B}_u[r], \sigma_u[r]$ to $u_0$}
		}
		$w \leftarrow \argmax_{u\in\mathcal{U}^\textsf{ex}}\{\mathcal{B}_u[r]\}$\;
		$\mathcal{M}_{u_0} \leftarrow \mathcal{M}_{u_0} \cup\{\sigma_w[r]\}$\;
		$\mathcal{M}_w\leftarrow \mathcal{M}_w\oplus \sigma_w[r]$\hfill\tcp{send $\mathcal{M}_w[r]$ to $w$}
	}
	$\mathcal{M}_{u_0}\leftarrow \texttt{solve}(\overline{P[u_0|\mathcal{M}_{u_0}]})$\;
	\KwRet{$\bigcup_{u\in\mathcal{U}}{\mathcal{M}_u}$}
\end{algorithm}

\DontPrintSemicolon
\SetKwProg{Fn}{Function}{}{}
\SetAlgoVlined
\SetKwFor{ForEach}{for each}{do}{endfch}
\begin{algorithm}
  	\caption{\textsf{cbba} EOSCSP solver} \label{alg:cbba}
  	\KwData{An EOSCSP $P = \langle\mathcal{S}, \mathcal{U}, \mathcal{R}, \mathcal{O}\rangle$}
  	\KwResult{An assignment $\mathcal{M}$} %
	$\mathcal{M}_{u_0}\leftarrow \emptyset$\;
	\SetKwFor{ForEach}{for each}{do concurrently}{endfch}
	\lForEach{$u\in\mathcal{U}^\textsf{ex}$}{$\mathcal{M}_u\leftarrow \texttt{solve}(P[u])$}
	\SetKwFor{ForEach}{for each}{do}{endfch}
	\ForEach{$r\in\texttt{sorted}(\mathcal{R})$}{
		\ForEach{$u\in\mathcal{U}^\textsf{ex}$}{
			$\mathcal{N}_u\leftarrow\texttt{candidates}(r)$\;
			$\mathcal{R}_u\leftarrow\mathcal{R}_u\cup\{r\}$
		}
	}
	\While{conflict}{
		\SetKwFor{ForEach}{for each}{do concurrently}{endfch}
		\ForEach{$u\in\mathcal{U}^\textsf{ex}$}{
			$\mathcal{B}_u, \mathcal{W}_u, \mathcal{T}_u\leftarrow \texttt{bundle}(u)$\;
			\tcp{send $\mathcal{B}_u, \mathcal{W}_u, \mathcal{T}_u$ to $\mathcal{N}_u$}
		}
		\ForEach{$u\in\mathcal{U}^\textsf{ex}$}{
			\tcp{solve conflicts and determine $\mathcal{M}_u$ (see \cite{Choi2009})}
		}
	}
	\ForEach{$u\in\mathcal{U}^\textsf{ex}$}{
		$\mathcal{M}_{u_0}\leftarrow\mathcal{M}_{u_0}\cup\{(o, t)|(o, t)\in \mathcal{M}_u, u_o = u_0\}$
	}
		
	$\mathcal{M}_{u_0}\leftarrow \texttt{solve}(\overline{P[u_0|\mathcal{M}]})$\;
	\KwRet{$\bigcup_{u\in\mathcal{U}}{\mathcal{M}_u}$}
\end{algorithm}

In PSI and SSI, the $\oplus$ operator is used to add $\sigma_u[r] = (o, t)$ in the current plan. Depending on the setting, it can be a simple aggregation if there is no conflict, or may require removing some already planned observations with lower reward. In SSI and CBBA, requests are sorted before looping over. This sorting can be done wrt due date, reward, or any combination of criteria. In the experiments, we will use the due date.

%% file: dcop.tex
Another approach to implement the allocation of requests between the multiple candidate exclusive users is to adopt a distributed constraint optimization vision. We devise here a cooperation mechanism between exclusive users to coordinate their scheduling process, by exchanging messages to reach an agreement on request allocations while meeting the coupling constraints, such as the capacity constraints.

\subsection{Some Background on DCOP}

One way to model inter-agent coordination problems is to
formalize them as distributed constraint optimization problems
(DCOP) \cite{Petcu2005}. 

\begin{definition}A discrete \emph{Distributed Constraint Optimization Problem} (or
  DCOP) is a tuple
  $\langle \mathcal{A}, \mathcal{X}, \mathcal{D}, \mathcal{C}, \mu
  \rangle$, where:
  \begin{inparaitem}[]
  \item[] $\mathcal{A} = \{a_1,\ldots,a_{|A|}\}$ is a set of agents;
  \item[] $\mathcal{X} = \{x_1,\ldots, x_n\}$ are variables owned by
    the agents;
  \item[]
    $\mathcal{D} = \{\mathcal{D}_{x_1},\ldots,\mathcal{D}_{x_n}\}$ is
    a set of finite domains, such that variable $x_i$ takes values in
    $\mathcal{D}_{x_i} = \{v_1,\ldots, v_k\}$;
  \item[] $\mathcal{C} = \{c_1,\ldots,c_m\}$ is a set of soft
    constraints, where each $c_i$ defines a cost
    $\in \mathbb{R}^+ \cup \{+\infty\}$ for each combination of
    assignments to a subset of variables (a constraint is initially
    known only to the agents involved);
  \item[] $\mu: \mathcal{X} \rightarrow \mathcal{A}$ is a function
    mapping variables to their associated agent;
  \item $f : \prod\mathcal{D}_{x_i} \to \mathbb{R}$ is an objective
    function, representing the global cost of a complete variable
    assignment.
  \end{inparaitem}\label{def:dcop}
The optimization objective is represented by function $f$, which, in
general, is considered as the sum of costs: $f = \sum_i c_i$.
  A \emph{solution} to a DCOP $P$ is a complete assignment to all
  variables. A solution is \emph{optimal} if it minimizes~$f$.
\end{definition}
  
DCOP have been widely
studied and applied in many areas of reference \cite{Fioretto2018}. They have many
interesting properties:
\begin{inparaenum}[(i)]
\item focus on decentralized approaches where agents negotiate a joint
  solution through local message exchanges;
\item exploitation of the domain structure (by encoding it in
  constraints) to address hard computational problems;
\item wide variety of solution methods ranging from exact methods to
  heuristic and approximate techniques; such as, for example, ADOPT
  \cite{Modi2005}, DPOP \cite{Petcu2005}, MaxSum
  \cite{Farinelli2008}, DSA \cite{Zhang2005} or MGM
  \cite{Maheswaran2004}, to name only the most famous.
\end{inparaenum} %

\subsection{Coordinating Exclusive Users with DCOP}

As for combinatorial auctions, using DCOPs for allocating all requests as a whole is too computationally expensive to be used. So we will use the same idea than SSI, and consider requests sequentially, and coordinate exclusive user for each request using a DCOP solver, instead of auctions in SSI.
This lets the non exclusive users coordinate to choose which one will fulfill it by scheduling an observation in its exclusive time windows.
Algorithm~\ref{alg:itnex2exdcop} sketches this method, coined \textsf{s\_dcop} (for sequential DCOP).
First, exclusive users also solve their own local sub-problem concurrently (line 1). Then, for each request $r$ in the ordered list of remaining requests (line 2), a new DCOP instance is collectively built between the exclusive users (line 3), and then solved (line 4) using any DCOP solver available (DPOP in our experiments). Once all requests have been considered, $u_0$ gathers the sub-solutions to build its own final solution, by scheduling as many observation outside exclusive time windows as possible (line 6-7). Note that, the inner plan of each exclusive user remains private, and only non exclusive schedule observation are communicated to $u_0$ (plus some extra information to handle inter-observation transition times).

\DontPrintSemicolon
\SetKwProg{Fn}{Function}{}{}
\SetAlgoVlined
\SetKwFor{ForEach}{for each}{do}{endfch}
\vspace{-0.6em}
\begin{algorithm}
  	\caption{\textsf{s\_dcop} EOSCSP solver} \label{alg:itnex2exdcop}
  	\KwData{An EOSCSP $P = \langle\mathcal{S}, \mathcal{U}, \mathcal{R}, \mathcal{O}\rangle$}
  	\KwResult{An assignment $\mathcal{M}$} %
	\SetKwFor{ForEach}{for each}{do concurrently}{endfch}
	\lForEach{$u\in\mathcal{U}^\textsf{ex}$}{$\mathcal{M}_u\leftarrow \texttt{solve}(P[u])$}
	\SetKwFor{ForEach}{for each}{do}{endfch}
	\ForEach{$r\in\texttt{sort}(\mathcal{R})$}{
		$p\leftarrow\texttt{build\_DCOP}(\theta_r, \mathcal{M}, \mathcal{M}_{u_1}, \ldots, \mathcal{M}_{u_n}, P)$\;
		$\mathcal{M}_{u_1}, \ldots, \mathcal{M}_{u_n} \leftarrow\texttt{solve\_DCOP}(p)$\;
		\SetKwFor{ForEach}{for each}{do concurrently}{endfch}
		\ForEach{$u\in\mathcal{U}^\textsf{ex}$}{
			$\mathcal{M}_u' \leftarrow \{(o,t) \in \mathcal{M}_u | u_o\in\mathcal{U}^\mathsf{nex}\}$\;
			\tcp{send $\mathcal{M}_u'$ to $u_0$}
			}
	}
	$\mathcal{M}_{u_0}\leftarrow \texttt{solve}(\overline{P[u_0|\displaystyle\bigcup_{u\in\mathcal{U}^\textsf{ex}}\mathcal{M}'_u]})$\;
	\KwRet{$\bigcup_{u\in\mathcal{U}}{\mathcal{M}_u}$}
\end{algorithm}

\subsection{DCOP Model}
Let's specify now the DCOP instance to be built in line 3 of Algorithm~\ref{alg:itnex2exdcop} for a given request $r$, and a current scheduling ($\mathcal{M}, \mathcal{M}_{u_1}, \ldots, \mathcal{M}_{u_n}$), as required in Definition~\ref{def:dcop}. Straightforwardly, the set of agents is the set of exclusive users which can potentially schedule the current request $r$: \begin{multline}\mathcal{A} = \{u\in\mathcal{U}^\textsf{ex}|\exists (s,(t_u^\text{start}, t_u^\text{end}))\in e_u,\exists o\in\theta_r\\\text{ s.t. } s_o=s, [t_u^\text{start}, t_u^\text{end}] \cap [t_o^\text{start}, t_o^\text{end}] \neq\emptyset \}\end{multline} 
We note $\mathcal{O}[u]^r = \{o\in\theta{r}|  \exists (s,(t_u^\text{start}, t_u^\text{end}))\in e_u,\text{ s.t. } s_o=s, [t_u^\text{start}, t_u^\text{end}] \cap [t_o^\text{start}, t_o^\text{end}] \neq\emptyset\}$ these observations related to request $r$ that can be scheduled on $u$'s exclusives.
Each such agent will own binary decision variables, one for each observation $o\in\mathcal{O}[u]^r$ and exclusive $e$ in its exclusives $e_u$, stating whether it schedules $o$ in $e$ or not: 
\begin{align}
\textstyle\mathcal{X} = \{x_{e,o}|e\in\bigcup_{u\in\mathcal{A}}e_u, o\in\mathcal{O}[u]^r\}\\
\mathcal{D} = \{\mathcal{D}_{x_{e,o}} = \{0, 1\}|x_{e,o}\in\mathcal{X}\}
\end{align}

The mapping $\mu$ associates each variable $x_{e,o}$ to $e$'s owner.

Constraints should check that at most one observation is scheduled per request (\ref{eq:amo_obs}), that satellites are not overloaded (\ref{eq:capa}), that at most one agent serves the same observation (\ref{eq:overlap}).
\begin{align}
\textstyle\sum_{e\in\bigcup_{u\in\mathcal{A}}e_u}x_{e,o}\leq 1,\quad \forall u\in\mathcal{X}, \forall o \in\mathcal{O}[u]^r\label{eq:amo_obs}\\
\hspace{-1em}\textstyle\sum_{o\in\{o\in\mathcal{O}[u]^r|u\in\mathcal{A},s_o=s\}, e\in\bigcup_{u\in\mathcal{A}}e_u}x_{e,o}\leq \kappa_s^*, \  \forall s\in\mathcal{S}
\label{eq:capa}
\end{align}
with $\kappa_s^*$ being the current capacity of $s$ given the already scheduled observations in $\mathcal{M}, \mathcal{M}_{u_1}, \ldots, \mathcal{M}_{u_n}$.
\begin{equation}
\label{eq:overlap}
\textstyle\sum_{e\in\bigcup_{u\in\mathcal{A}}e_u}x_{e,o} \leq 1,\quad\forall o\in\mathcal{O}
\end{equation}

Beside, the cost to integrate an observation in the current user's schedule should be assessed to guide the optimization process. We thus add soft constraint to each $x_{e,o}$:
\begin{equation}
c(x_{e,o}) = \pi(o, \mathcal{M}_{u_o}), \quad\forall x_{e,o} \in\mathcal{X}\label{eq:pi}
\end{equation}
where $\pi$ evaluates the best cost obtained when scheduling $o$ and any combination of observations from $\mathcal{M}_{u_o}$, as to consider all possible revisions of $u_o$'s current schedule. Practically, instead of computing $\pi$ each time, some constraint compilation can be used to assess all these combinations only once. These exponential number of alternatives are evaluated using polynomial greedy algorithm.
To sum up:
\begin{equation}
\mathcal{C} = \{(\ref{eq:amo_obs}),(\ref{eq:capa}),(\ref{eq:overlap}),(\ref{eq:pi})\}
\end{equation}

%% file: experiments.tex
Experiments aim to analyze the performances of the investigated algorithms with a growing number of requests (and observations). They are coded in Python 3.7 and executed on 20-core Intel(R) Xeon(R) CPU E5-2660 v3 @ 2.60GHz, 62GB RAM, Ubuntu 18.04.5 LTS. We ran 30 instances of randomly generated EOSCSP with seed in [0:29] for each problem size, and plot the average, with [0.05, 0.95] confidence. The \texttt{solve} procedure used in \textsf{psi}, \textsf{ssi}, \textsf{cbba} and \textsf{s\_dcop} is the greedy algorithm. The DCOP algorithm used by \textsf{s\_dcop} is the DPOP implementation from pyDCOP \cite{Rust2019}. Randomly generated values are uniformly chosen within provided intervals. The computation time reported latter is a centralized computation time (no real distribution over several computers).

\begin{figure*}[t]
\centering
     \begin{subfigure}[b]{0.245\textwidth}
         \centering
         \includegraphics[width=\textwidth]{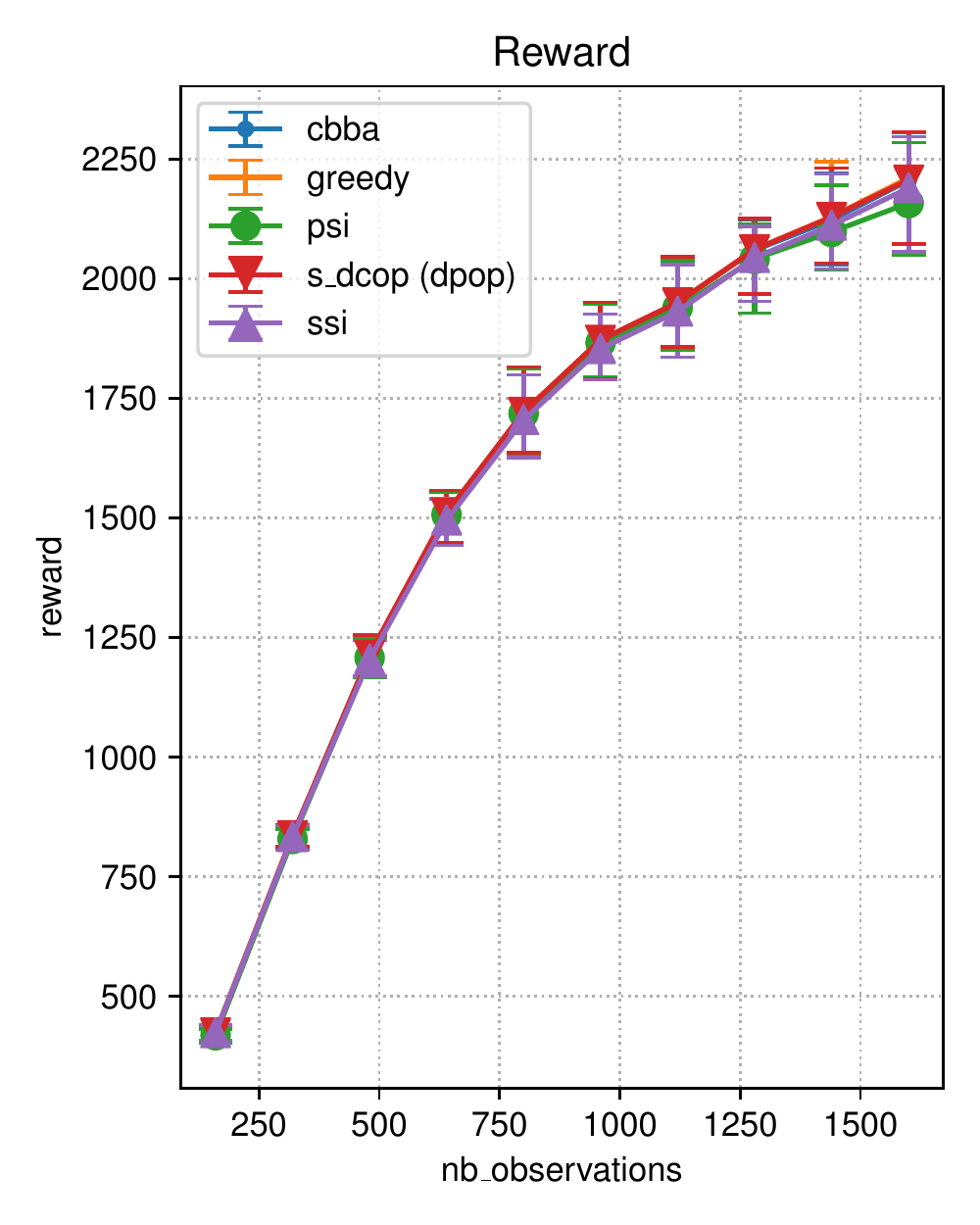}
         \label{fig:stressed_reward}
     \end{subfigure}
     \hfill
     \begin{subfigure}[b]{0.245\textwidth}
         \centering
         \includegraphics[width=\textwidth]{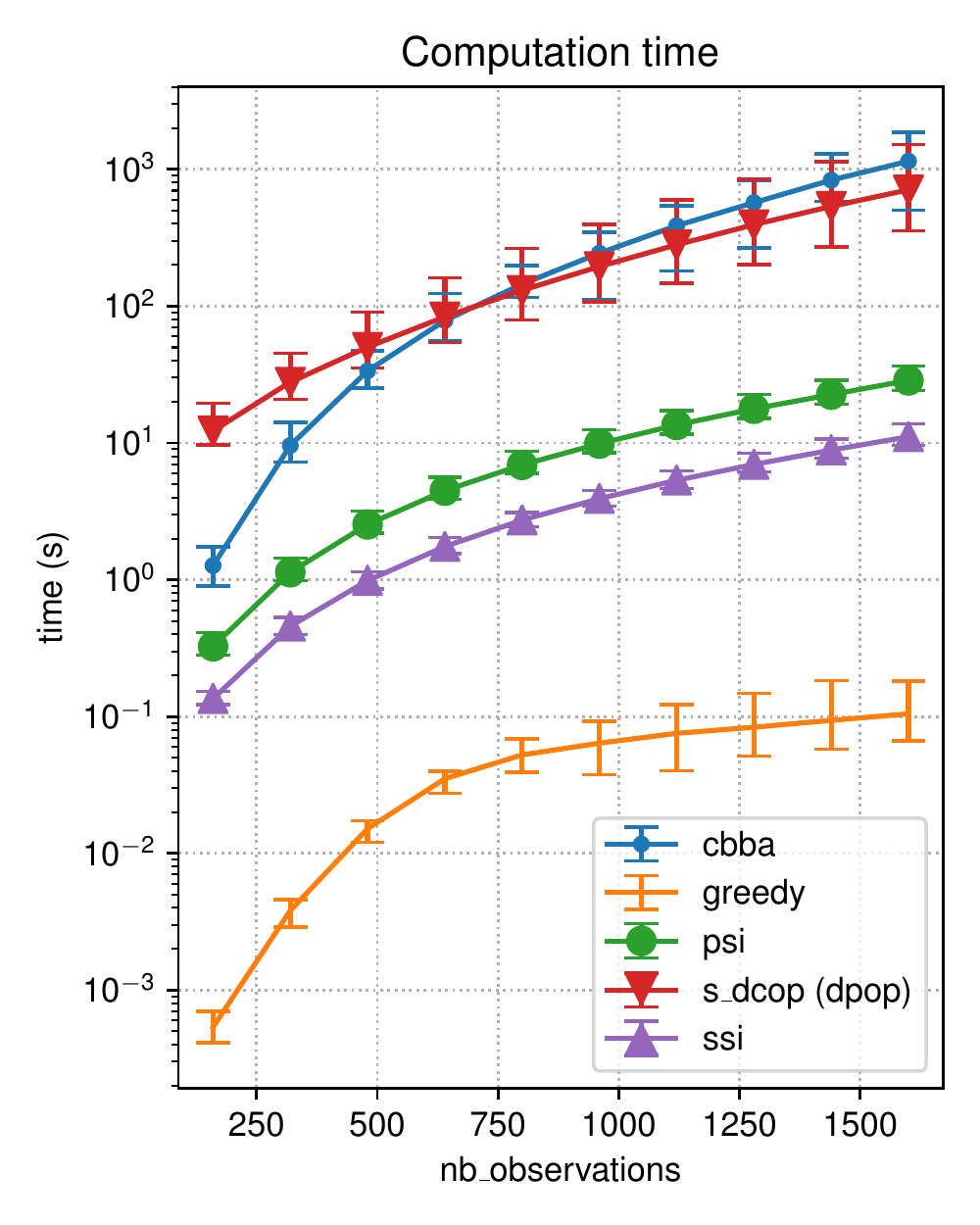}
         \label{fig:stressed_time}
     \end{subfigure}
     \hfill
     \begin{subfigure}[b]{0.245\textwidth}
         \centering
         \includegraphics[width=\textwidth]{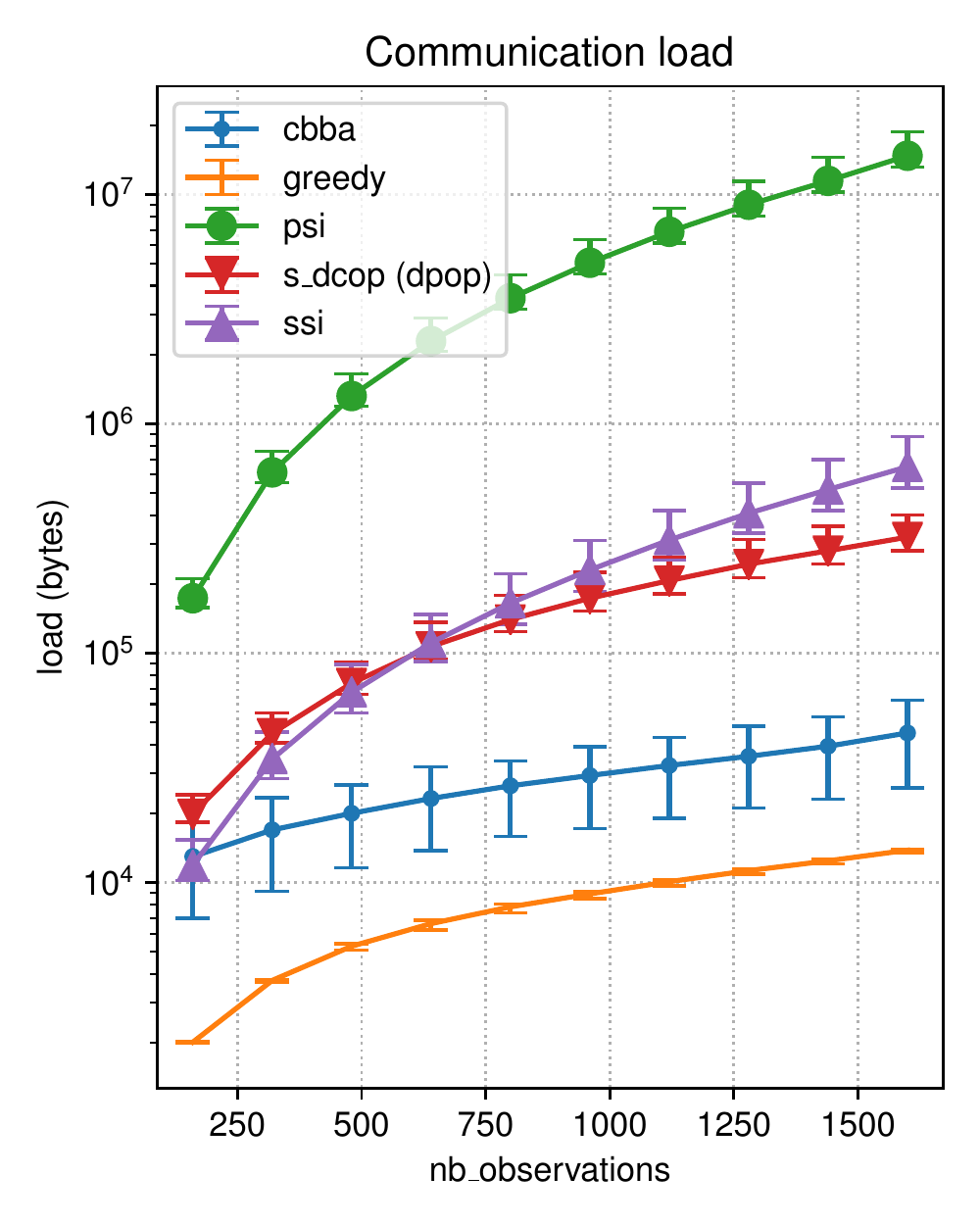}
         \label{fig:stressed_size_messages}
     \end{subfigure}
     \hfill
     \begin{subfigure}[b]{0.245\textwidth}
         \centering
         \includegraphics[width=\textwidth]{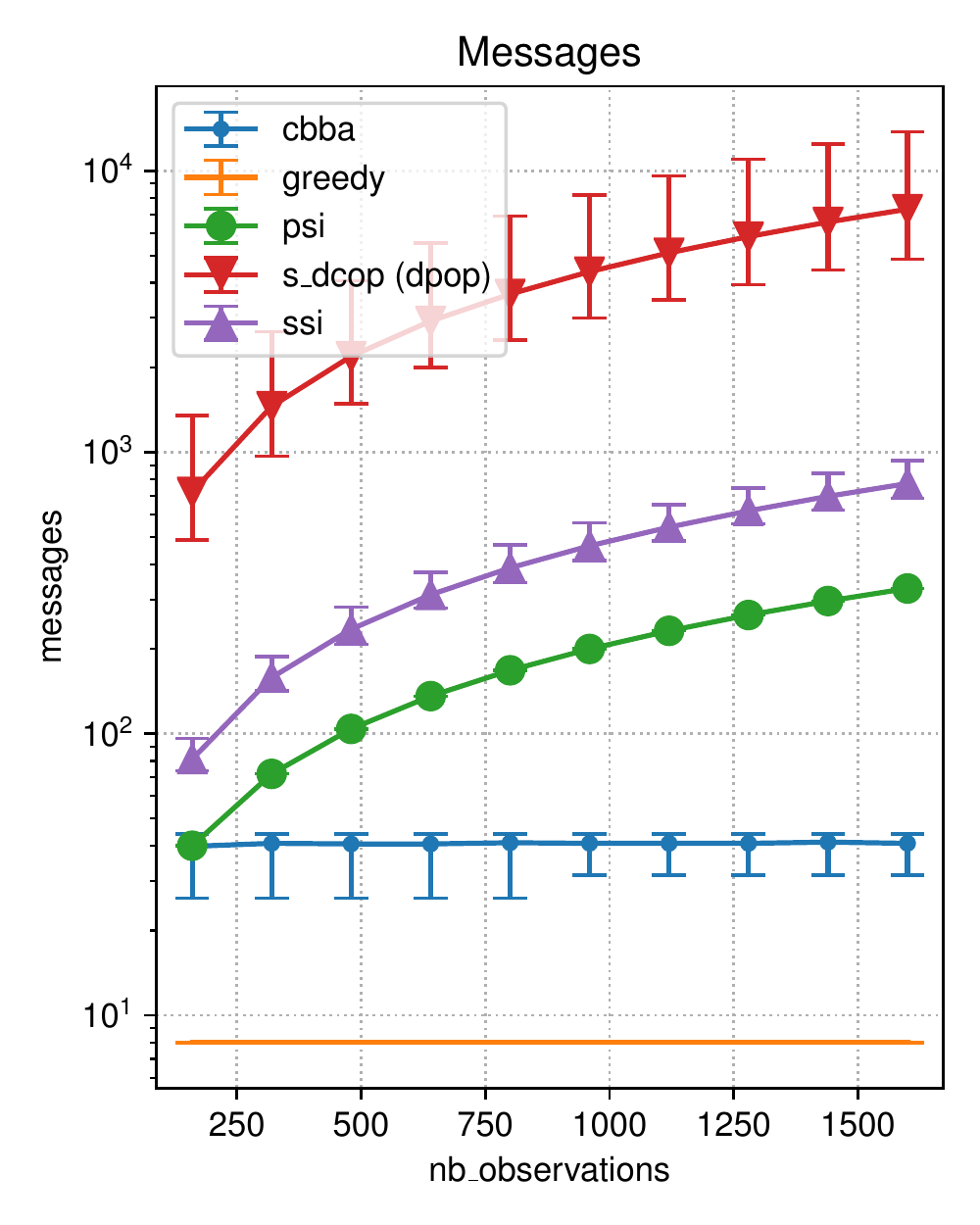}
         \label{fig:stressed_nb_messages}
     \end{subfigure}
     \vspace{-2em}
        \caption{Results for the investigated distributed solution methods on highly conflicting small-scale problems.}
        \label{fig:stressed}
\end{figure*}

\begin{figure*}[t]
\centering
     \begin{subfigure}[b]{0.245\textwidth}
         \centering
         \includegraphics[width=\textwidth]{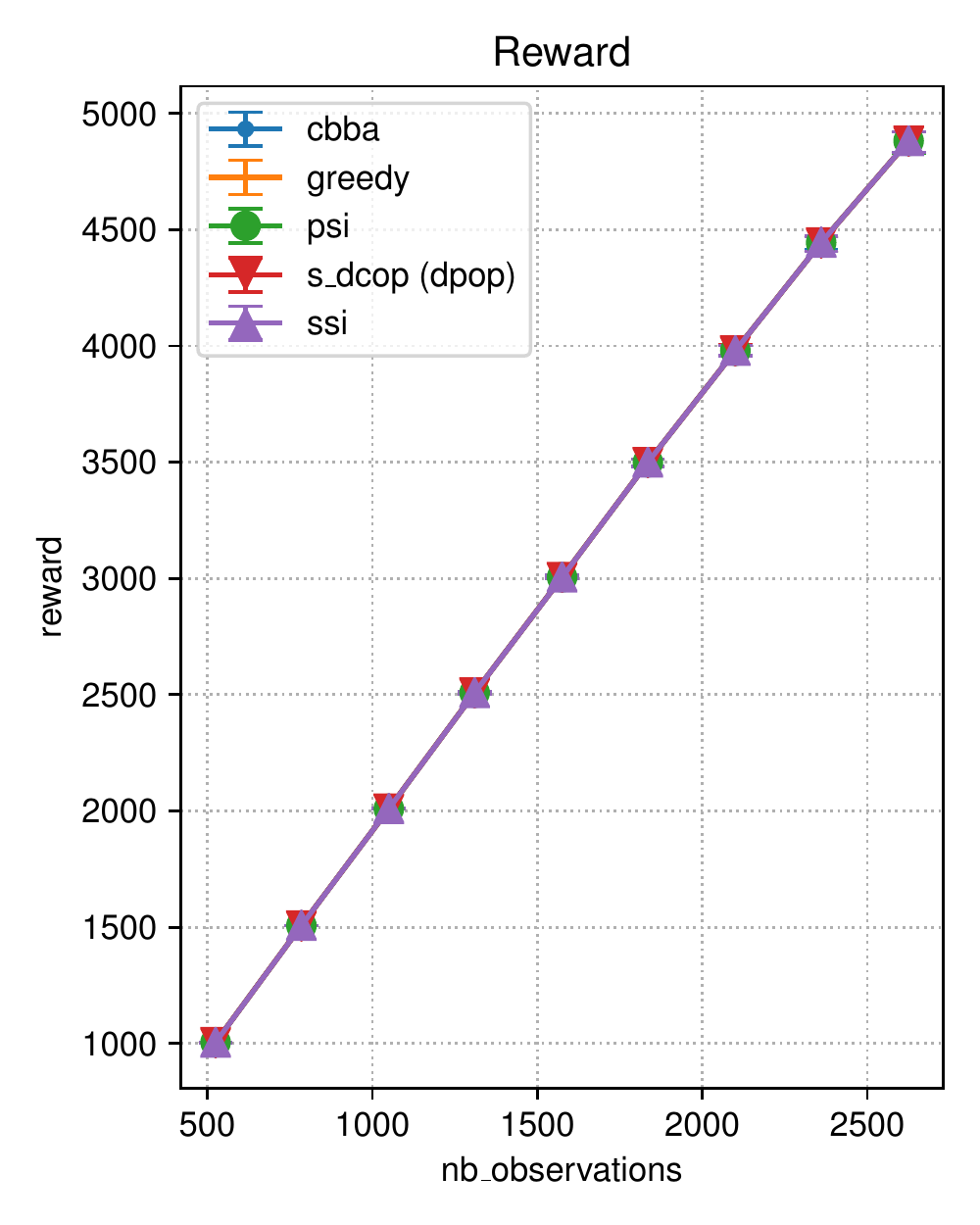}
         \label{fig:stressed_reward_r}
     \end{subfigure}
     \hfill
     \begin{subfigure}[b]{0.245\textwidth}
         \centering
         \includegraphics[width=\textwidth]{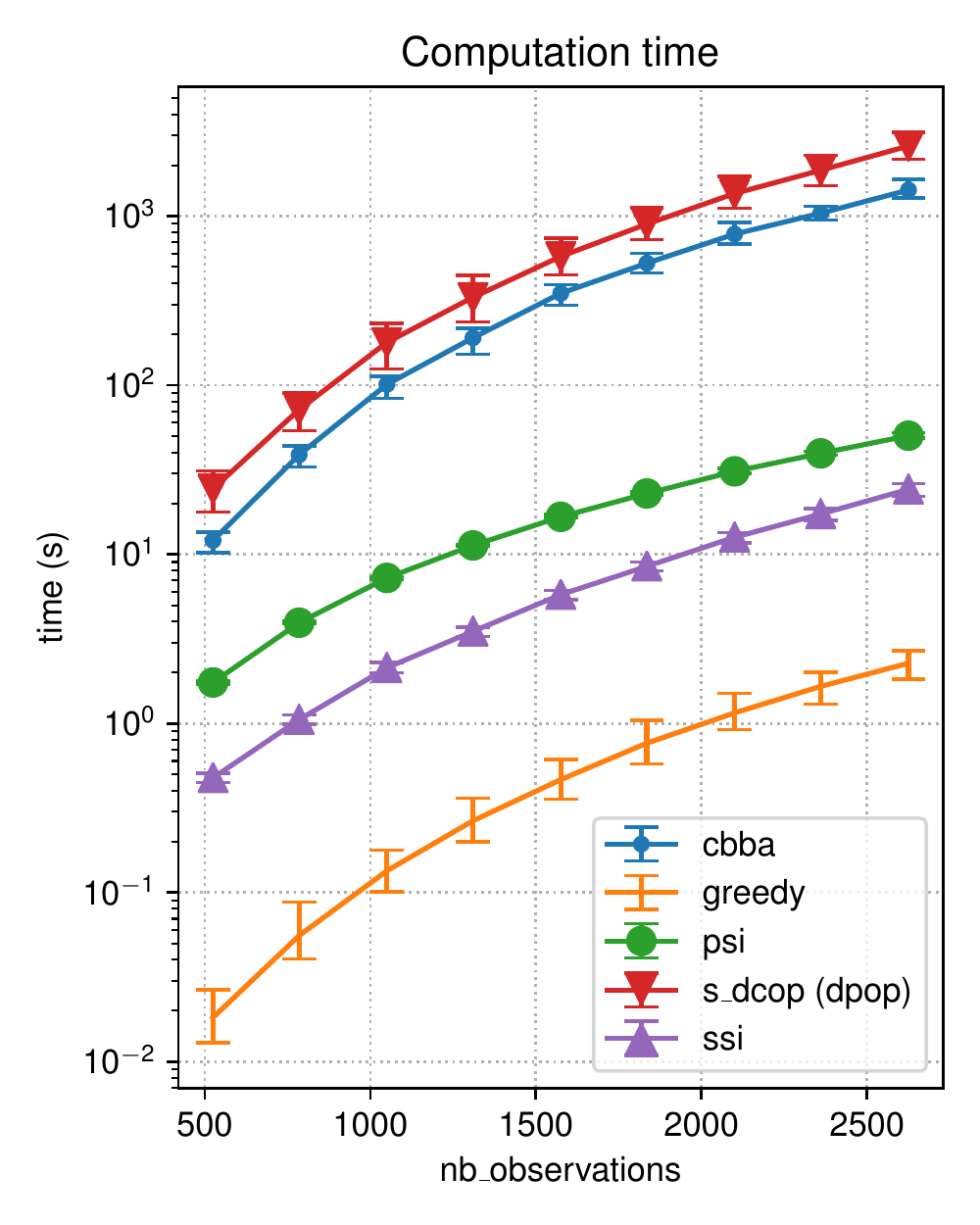}
         \label{fig:stressed_time_r}
     \end{subfigure}
     \hfill
     \begin{subfigure}[b]{0.245\textwidth}
         \centering
         \includegraphics[width=\textwidth]{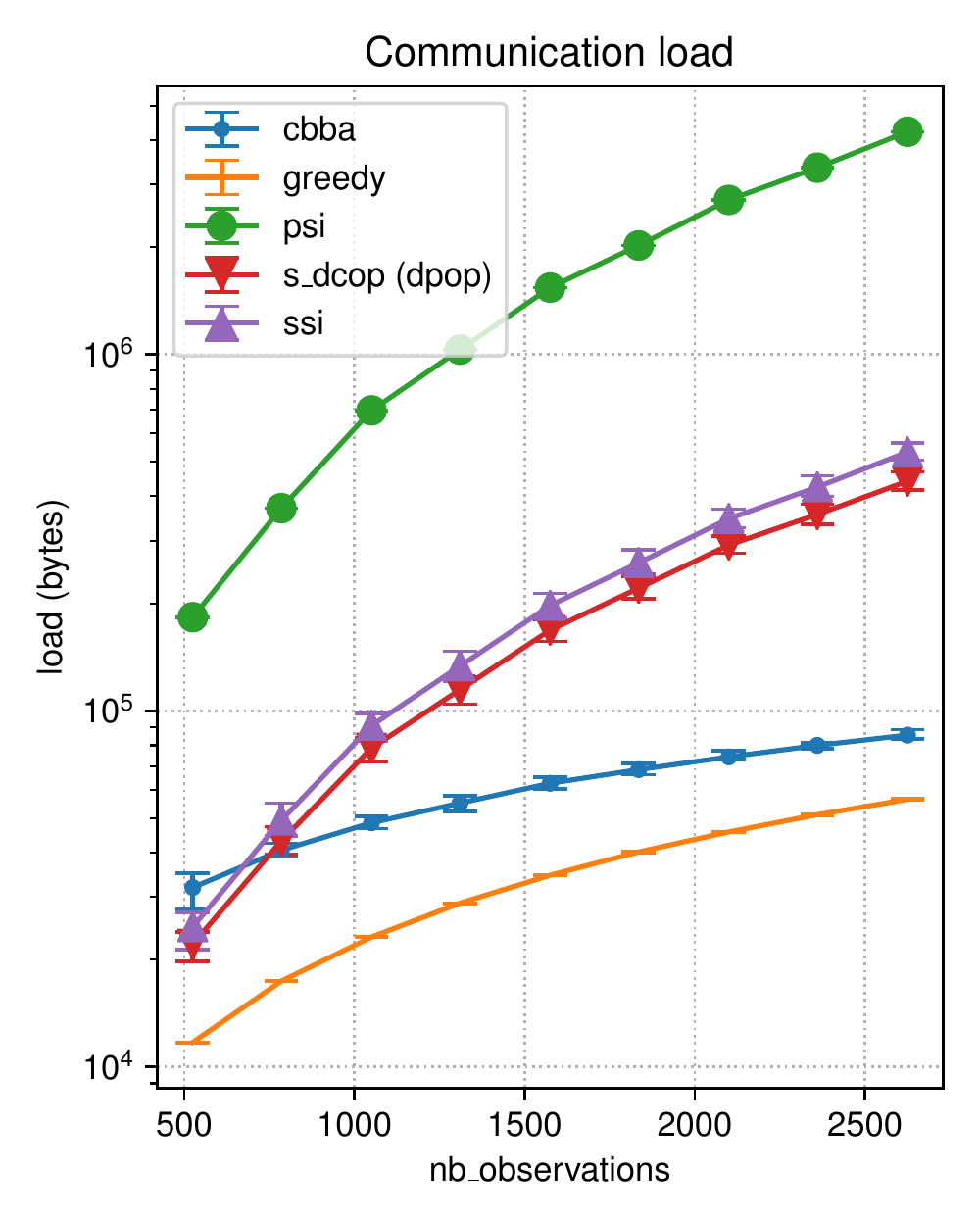}
         \label{fig:stressed_size_messages_r}
     \end{subfigure}
     \hfill
     \begin{subfigure}[b]{0.245\textwidth}
         \centering
         \includegraphics[width=\textwidth]{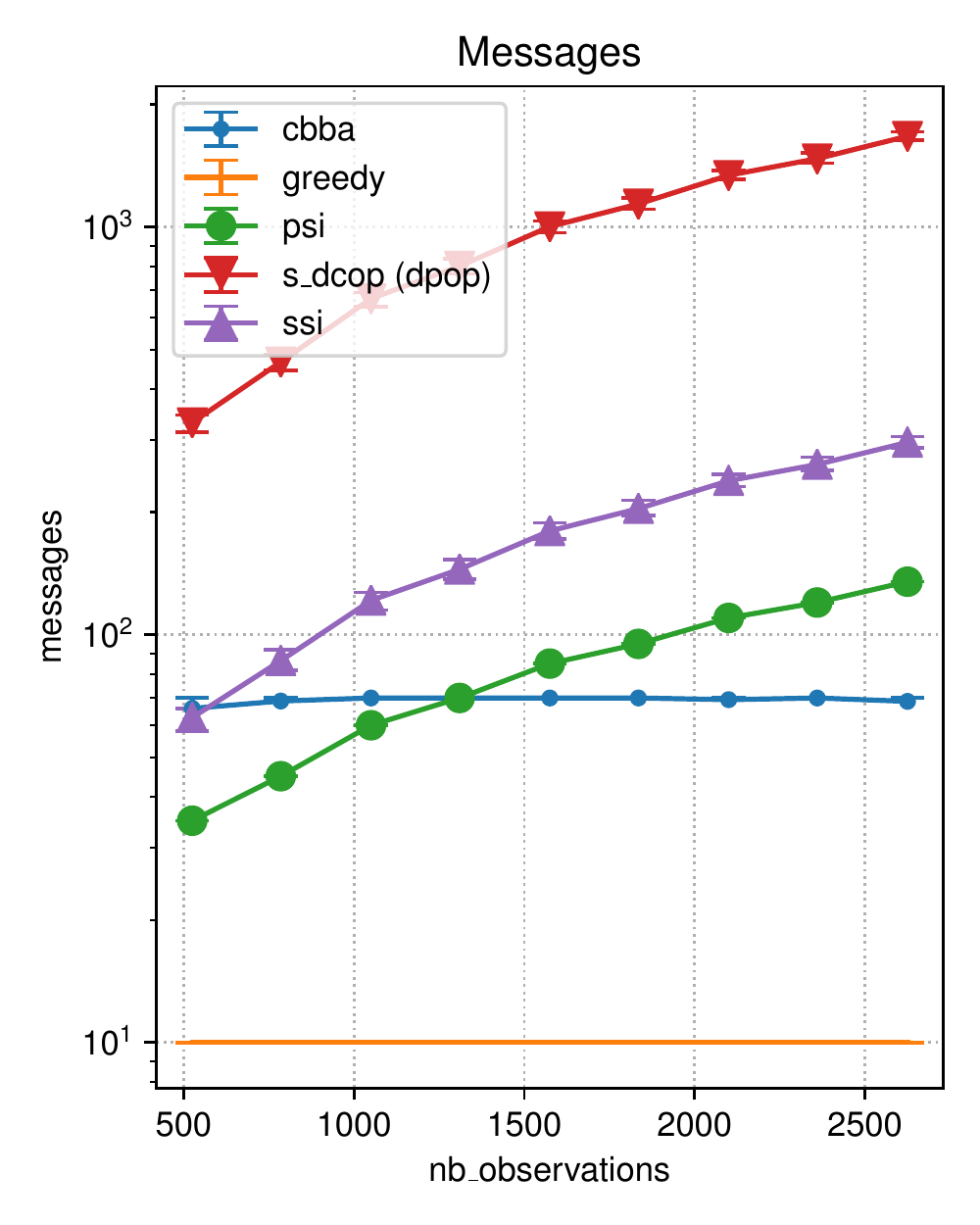}
         \label{fig:stressed_nb_messages_r}
     \end{subfigure}
     \vspace{-2em}
        \caption{Results for the investigated distributed solution methods on problems with realistic large-scale order books.}
        \label{fig:realistic}
\end{figure*}

\paragraph{Highly conflicting problems.} We evaluate the algorithms on very conflicting small-scale problems (5 min planning horizon). We generate EOSCSPs with 3 satellites with a capacity of 20 observations, 4 exclusive users emitting 2 to 20 requests each, 8 exclusive portions per user with a random duration in [15:20], a central planner emitting 8 to 80 requests, 10 observation opportunities per request of duration 5 that can be scheduled in a time window with duration in [10:20], and a reward in [10:50:10] for exclusive user, and in [1:5] for central planner. Satellites' time window is [0, 300]. Transition times between observations are uniformly equals to 1. Exclusives are randomly positioned, while ensuring they do not overlap. Observation time windows are randomly positioned, as to ensure they are either included in one exclusive, or outside any exclusive. There are many observation overlaps, and as many requests from central planner than all requests from the exclusive users. 

Figure~\ref{fig:stressed} shows the results for this setting. Reward-wise, all the distributed algorithms except \textsf{itnex2ex} are almost as good as \textsf{greedy}, which is our baseline. Still, \textsf{cbba} and \textsf{s\_dcop} provides the best distributed solutions. The decline with growing number of observations is due to the satellites' capacity saturation. \textsf{s\_dcop} and \textsf{cbba}'s performances are at the cost of extra computation time, while remaining reasonable (approx. 1000 seconds), contrary to optimal solver (e.g. CPLEX) that cannot solve instances with more than $100$ observations (not displayed here). \textsf{s\_dcop}'s higher computation time results from pre-computing function $\pi$ and the underlying DPOP solving procedure. \textsf{cbba} computational overhead is due to bundle valuation. At some point (problems larger than 750 observations to schedule), \textsf{cbba} requires more time to compute than \textsf{s\_dcop}. This is due to the exponentially growing number of bundles to consider and the fact that at this size, with such a conflicting setting, the \textsf{cbba} neighborhood network is a complete graph, meaning that each user has to resolve conflicts with all the other users. Communication-wise, \textsf{psi} exchanges few large messages, since all requests are communicated to all users, resulting in exchanging more 10Mb in larger instances. \textsf{ssi} and \textsf{s\_dcop} exchange numerous messages of smaller size (only sending bids on requests of interest), due to the sequential process they follow. On its side, \textsf{cbba} exchanges fewer messages of small size (approx. total 30kB in large instances), which makes it a very relevant candidate in distributed settings, with good compromise between solution quality and communication load. If reactiveness is a requirement, \textsf{ssi} remains the best candidate.

\paragraph{Realistic problems.} Here, we generate large-scale EOSCSPs, with realistic parameters, with respect with order books provided by our partners, to schedule thousands of observations in a 6-hour planning horizon. We generate instances as previously but with 8 satellites with a capacity of 500 observations, 5 exclusive users with 20 to 100 requests each, 10 exclusive orbit portions per user with a duration in [300:600], 1 central planner with 25 to 250 requests, 5 observation opportunities per request of duration of 20 that can be scheduled in a time window with duration in [40:60] included in an exclusive windows (there is no request outside exclusive windows in this setting), and the planning time window is [0, 21600].

Figure~\ref{fig:realistic} shows results for this setting. All algorithms provide good quality solutions equivalent to \textsf{greedy}. The results obtained  in this setting, only focusing on observations inside exclusive windows, confirm the performances of the different benchmarked algorithm, except that here  \textsf{cbba} does require more time to compute than \textsf{s\_dcop}. This is due to the fact that these larger instances are less conflicting, and that the neighborhoods are no longer complete. Let's note that both \textsf{s\_dcop} and \textsf{cbba} are very distributed in nature, and performs many computation concurrently. Therefore there is room for computation speedup in real distributed settings.

%% file: conclusion.tex
This paper investigated for the first time the use of distributed and multi-agent techniques to solve the novel EOSCSP, keeping in mind the need to limit information disclosure between users. We defined core components of EOSCSP, and proposed a straightforward MILP encoding to optimally solve such problems. This is unfortunately non usable in practice, even on small instances. We thus proposed a greedy and fast algorithm to solve EOSCSP. We devised and implemented several distributed algorithms (\textsf{psi}, \textsf{ssi}, \textsf{cbba} and \textsf{s\_dcop}), all keeping the inner user plans private. \textsf{s\_dcop} and \textsf{cbba} provides solutions equivalent to the best evaluated algorithms on over-conflicting problems. This has a cost: higher communication load and computation time to assess the reward to integrate an observation in a given schedule. Yet, these techniques are fully distributable, and may gain from concurrent execution. On realistic large scale problems, the solution quality is still very good wrt. \textsf{greedy}. While, these problems still require less coordination because the probability for overlapping observations is smaller, EOSCSP still implies numerous observations from exclusive users, which makes the computation of the \textsf{s\_dcop} evaluation function $\pi$ and the construction of the \textsf{cbba} bundles expensive. A good compromise is thus to use \textsf{ssi} in larger settings, since computation time, communication-load are very limited, while providing good quality solutions. Note that this investigation was also a very good terrain for confronting DCOP-based and Market-based techniques, which are most often not compared in the literature.

This work raises several perspectives, notably the development of dedicated DCOP or CBBA solvers adapted to EOSCSP specificity, e.g. the use of the evaluation function $\pi$ or the construction of bundles, that may result from a learning process, instead of a systematic assessment of every alternative. On may also consider devising dedicated bidding language to assess bundles and perform the winner determination problem in a efficient manner. Finally, we are currently working on integrating uncertainties about observation success into the decision process, which leads to even more complex problems to solve. 